\documentclass{article}

    \usepackage[preprint]{neurips_2025}

\usepackage[utf8]{inputenc} % allow utf-8 input
\usepackage[T1]{fontenc}    % use 8-bit T1 fonts

\usepackage{xcolor}         % colors
\definecolor{FirebrickRed}{HTML}{B22222}
\definecolor{SteelBlue}{HTML}{4682B4}
\definecolor{PeruColor}{HTML}{CD853F}
\definecolor{TealColor}{HTML}{008080} % Using TealColor to avoid potential name clashes
\usepackage[colorlinks=true,
            linkcolor=TealColor,
            citecolor=SteelBlue,
            urlcolor=FirebrickRed,
            ]{hyperref}       % hyperlinks
            
\usepackage{url}            % simple URL typesetting
\usepackage{booktabs}       % professional-quality tables
\usepackage{amsfonts}       % blackboard math symbols
\usepackage{nicefrac}       % compact symbols for 1/2, etc.
\usepackage{microtype}      % microtypography
\usepackage{graphicx}       
\usepackage{amsmath}       

\usepackage[ruled,vlined]{algorithm2e}
\usepackage{wrapfig}        
\usepackage{algpseudocode}  
\usepackage{float}          
\usepackage{lipsum}         
\usepackage{booktabs}       
\usepackage{array}
\usepackage{multirow}
\usepackage{makecell}
\usepackage{graphicx}
\usepackage{enumitem}
\usepackage{caption}
\usepackage{gradient-text}
\def\NickName{{\gradientRGB{DeepVerse}{130,54,185}{19,127,241}}\xspace}
\usepackage{amssymb}
\usepackage{comment}
\usepackage{nicefrac}       % compact symbols for 1/2, etc.
\usepackage{microtype}      % microtypography
\usepackage[normalem]{ulem}

\title{\NickName: 4D Autoregressive Video Generation \\ as a World Model}

\usepackage{color}

\definecolor{url_color}{RGB}{42, 83, 163}
\author{%
    Junyi Chen\textsuperscript{1,2}\quad
    Haoyi Zhu\textsuperscript{2,3}\quad
    Xianglong He\textsuperscript{4}\quad
    Yifan Wang\textsuperscript{1,2${\spadesuit}$}\quad 
    Jianjun Zhou\textsuperscript{2,5${\spadesuit}$}\quad \\
    \textbf{Wenzheng Chang}\textsuperscript{2,3${\spadesuit}$}\quad 
    \textbf{Yang Zhou}\textsuperscript{2,6${\spadesuit}$}\quad 
    \textbf{Zizun Li}\textsuperscript{2,3${\spadesuit}$}\quad 
    \textbf{Zhoujie Fu}\textsuperscript{2,7} \\
    \textbf{Jiangmiao Pang}\textsuperscript{2}\quad 
    \textbf{Tong He}\textsuperscript{2${\heartsuit}$} \\
    \textsuperscript{1}SJTU \quad 
    \textsuperscript{2}Shanghai AI Lab \quad 
    \textsuperscript{3}USTC \quad 
    \textsuperscript{4}THU \quad 
    \textsuperscript{5}ZJU \quad 
    \textsuperscript{6}FDU \quad 
    \textsuperscript{7}NTU \\
    \textsuperscript{${\spadesuit}$}Equal contribution \quad 
    \textsuperscript{${\heartsuit}$}Corresponding Author \\
    \texttt{\url{https://sotamak1r.github.io/deepverse/}}
}

\begin{document}

\maketitle

\begin{figure}[h]
    \begin{center}
        \vspace{-1.5em}
        \includegraphics[width=\textwidth]{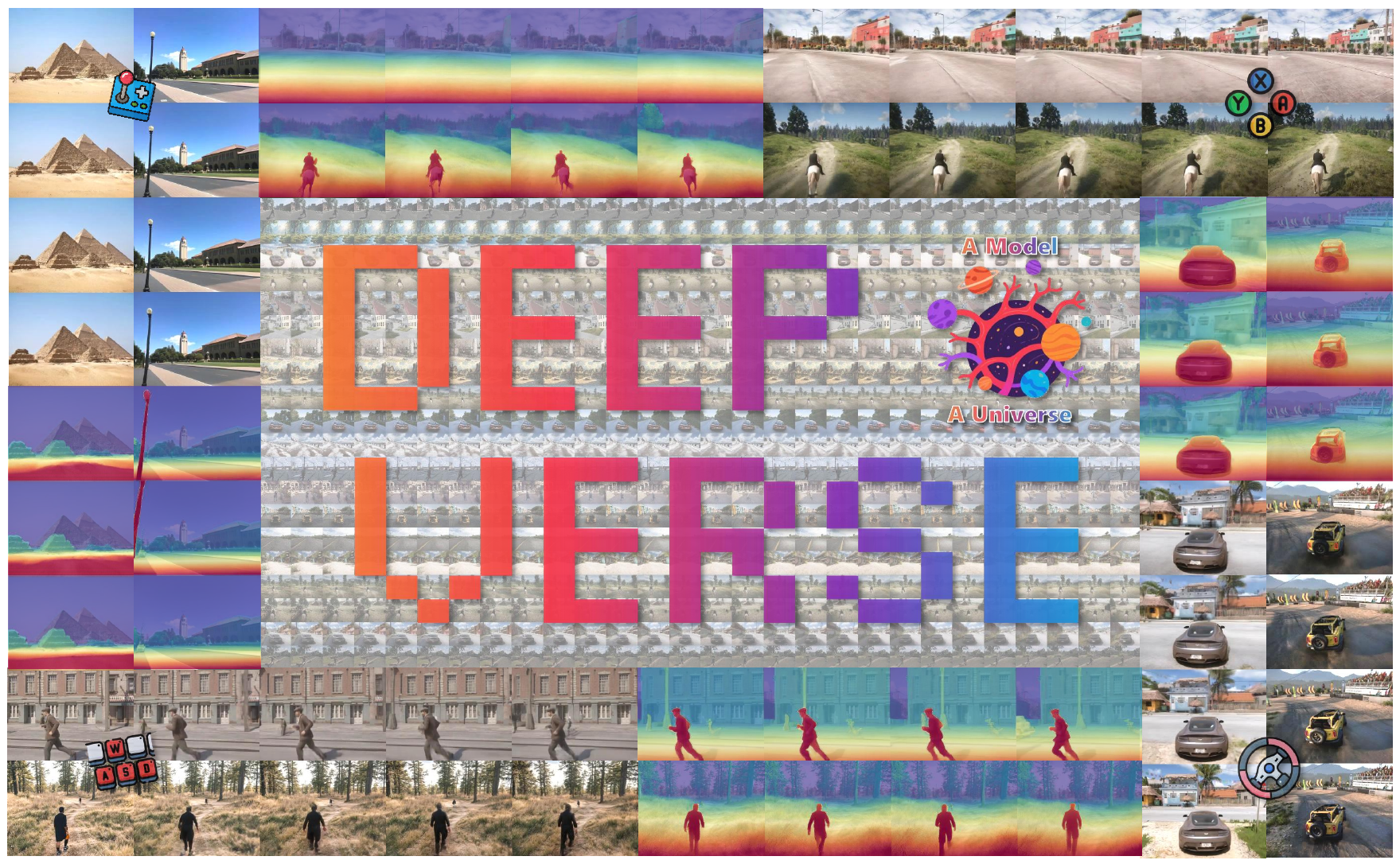}
        % \vspace{-1.5em}
        \caption{
            We introduce \NickName, an interactive world model grounded in 4D autoregressive video generation. 
            By establishing a 4D spatiotemporal distribution of the world, \NickName enables continuous and coherent 4D future prediction from merely a single input image, effectively modeling both spatial layouts and temporal dynamics simultaneously.
        }
        \label{fig:teaser}
    \end{center}
    \vspace{-1em}
\end{figure}

\begin{abstract}

World models serve as essential building blocks toward Artificial General Intelligence (AGI), enabling intelligent agents to predict future states and plan actions by simulating complex physical interactions.
However, 
% existing interactive models primarily rely on visual predictions, neglecting crucial geometric structures and spatial coherence, leading to rapid error accumulation and temporal inconsistency.
existing interactive models primarily predict visual observations, thereby neglecting crucial hidden states like geometric structures and spatial coherence. This leads to rapid error accumulation and temporal inconsistency.
To address these limitations, we introduce \NickName, a novel 4D interactive world model explicitly incorporating geometric predictions from previous timesteps into current predictions conditioned on actions. 
%This explicit geometric dependency significantly reduces drift and enhances temporal consistency, allowing our model to reliably generate extended future sequences.
%Experiments demonstrate that incorporating geometric constraints enables \NickName\space to capture richer spatial-temporal relationships and physical dynamics, substantially improving prediction accuracy, visual realism, and scene rationality. 
Experiments demonstrate that by incorporating explicit geometric constraints, \NickName\space captures richer spatio-temporal relationships and underlying physical dynamics. This capability significantly reduces drift and enhances temporal consistency, enabling the model to reliably generate extended future sequences and achieve substantial improvements in prediction accuracy, visual realism, and scene rationality.
Furthermore, our method provides an effective solution for geometry-aware memory retrieval, effectively preserving long-term spatial consistency. We validate the effectiveness of \NickName\space across diverse scenarios, establishing its capacity for high-fidelity, long-horizon predictions grounded in geometry-aware dynamics.

\end{abstract}

\section{Introduction}

Interactive understanding of the physical world is a fundamental task for intelligent systems.
World models, which 
aim to learn state transition functions from raw observations of external environments, provide essential predictive capabilities for intelligent agents, enabling them to imagine future states, evaluate possible actions, and navigate complex, dynamic scenarios.
Recent progress in world models has demonstrated considerable potential in tasks such as visual simulation~\cite{nvidia2025cosmosworldfoundationmodel}, embodied navigation~\cite{team2025aether}, and manipulation~\cite{zhen2025tesseract}.

Despite notable advancements in constructing effective world models, current online approaches~\cite{feng2024thematrix,valevski2024gamengen,oasis2024} still suffer significantly from cumulative prediction errors and the forgetting issue. 
Addressing the above challenges is non-trivial. 
Most existing methods~\cite{song2025historyguide, xiao2025worldmem} attempt to mitigate these issues by developing sophisticated techniques to efficiently incorporate historical frames. 
For instance, the recent FramePack~\cite{zhang2025framepack} compresses past frames into a fixed-length representation, thereby maintaining context within a transformer's limited memory. 
However, these visual-centric strategies fundamentally overlook a critical aspect: videos inherently represent 2D projections of a dynamic 3D/4D physical world. 
Without explicit modeling of underlying geometric structures, models inevitably struggle to maintain long-term accuracy and consistency in visual predictions.

To this end, we propose \NickName, the first autoregressive 4D world model trained on large-scale synthetic data with precise spatial labels. 
It explicitly incorporates geometric reasoning into online predictive modeling. 
By grounding visual forecasting in robust geometric representations, our method significantly enhances prediction accuracy, reduces drifting, and addresses the forgetting issue over extended temporal horizons. 
Specifically, our approach leverages a powerful autoregressive prior trained on large-scale real-world video data, capturing rich dynamic patterns and visual semantics. 
In parallel, we utilize extensive synthetic datasets that provide accurate ground-truth geometry supervision, including depth maps and camera poses. 

At each timestep, the model predicts the future state based not only on the previously generated RGB frames but also on the preceding geometric estimations.
In doing so, the model greatly mitigates inherent issue in purely visual autoregressive systems, such as the scale ambiguity (figure \ref{fig:scale_camera}a), addressing the core issues of drifting and forgetting in conventional methods.

Besides, to further address the forgetting issue, we introduce a geometry-aware memory read-and-write mechanism. 
We design a geometric memory module that compares the current geometry with historical observations and selects those with higher spatial overlap or structural similarity. 
These retrieved observations are then used as conditioning inputs for the current prediction step. This targeted memory retrieval allows the model to retain access to long-term contextual information without overwhelming the predictive pipeline, effectively reducing forgetting while maintaining spatial and temporal coherence over extended sequences.

To sum up, our contributions can be summarized as follows:

\begin{itemize}[left=0pt]
\vspace{-0.6em}
\item We present \NickName, the pioneering introduction of an autoregressive 4D world modeling paradigm, establishing theoretical and practical guidelines for architectural configuration selection in future interactive world model development.
\vspace{-0.6em}
\item We innovatively incorporate 4D information into an autoregressive world modeling framework. By explicitly constructing the 4D world during the generation process, the proposed methodology demonstrates significant enhancement in visual consistency while effectively addressing scale ambiguity issues inherent to unimodal visual paradigms.
\vspace{-0.6em}
\item Building upon the \NickName framework's capability for concurrent spatial distribution modeling, we have engineered a spatial memory mechanism to enhance long-term temporal consistency in generated sequences, thus establishing a robust framework for maintaining spatiotemporal continuity in autoregressive generation processes.
\end{itemize}

\section{Method}

To provide a comprehensive illustration of our method, 
we elaborate on our problem formulation in section \ref{sec:State Definition} 
and discribe the methodology for tailoring model architecture in section \ref{sec:Model Components}.
% Then we systematically address the construction of training datasets in section \ref{sec:Dataset Construction}. 
The construction of training datasets is systematically addressed in Section \ref{sec:Dataset Construction}. 
Finally, Section \ref{sec:Long-Duration Inference} demonstrates \NickName's operational workflow during the inference phase.

\begin{figure}[t]
    \begin{center}
    \vspace{-0.5em}
        \includegraphics[width=\textwidth]{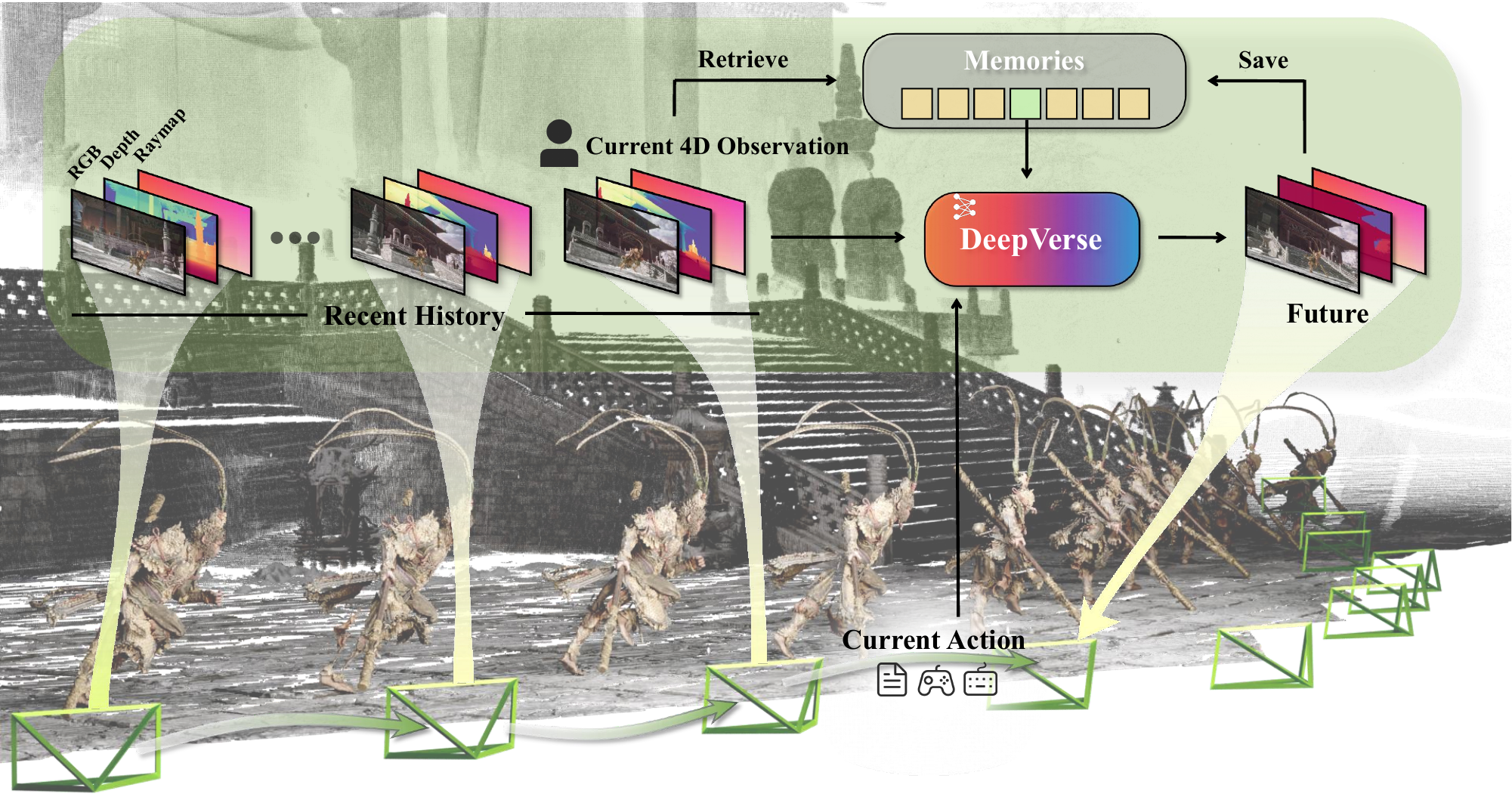}
        \caption{
            \textbf{Our framework.} 
            The inputs to \NickName consist of: 
            (1) a sequence of $m$ consecutive 4D observations encompassing current and recent estimated states; 
            (2) spatial conditions retrieved from a global memory pool through the selective mechanism $\psi$; 
            (3) textually specified control signals. 
            The system subsequently generates $k$ temporally coherent 4D future observations, which are automatically archived into the global memory repository for persistent world state tracking.
        }
        \label{fig:pipeline}
    \end{center}
    \vspace{-1.5em}
\end{figure}

%%%%%%%%%%%%%%%%%%%%%%%%%%%%%%%%%%%%%%%%%%%%%%%%%%%%%%%%%%%%%%%%%%%%%%%%%%%%%%%%%%%%%%%%%%%%%%%%%%%%%%%%%%%%%%%%
%%%%%%%%%%%%%%%%%%%%%%%%%%%%%%%%%%%%%%%%%%%%%%%%%%%%%%%%%%%%%%%%%%%%%%%%%%%%%%%%%%%%%%%%%%%%%%%%%%%%%%%%%%%%%%%%
%%%%%%%%%%%%%%%%%%%%%%%%%%%%%%%%%%%%%%%%%%%%%%%%%%%%%%%%%%%%%%%%%%%%%%%%%%%%%%%%%%%%%%%%%%%%%%%%%%%%%%%%%%%%%%%%
\subsection{Problem Formulation}
\label{sec:State Definition}

World models aim to learn the transition function $P(s_{t+1}|a_t, s_t)$ in a Markov Decision Process (MDP), where $s_t$ denotes the environment state at timestamp $t$. However, many real-world applications are Partially Observed MDPs (POMDPs) where $s_t$ is latent. While prior work~\cite{bruce2024genie,valevski2024gamengen,oasis2024} often used visual observations $v_t$ directly, these provide an incomplete, non-Markovian signal. To address this, \NickName introduces a \textit{composite 4D state representation} as a more informative proxy for $s_t$:
\begin{equation}
    \label{eq:4d_representation}
    \hat{s}_t = (v_t, g_t).
\end{equation}
Here, $v_t$ is the visual observation, and $g_t$ encapsulates geometric information, specifically relating to camera viewpoint $c_t$ and depth $d_t$. This $\hat{s}_t$ allows for a local 3D geometric representation, aiming to better approximate the hidden state $s_t$ than $v_t$ alone. A sequence of these $\hat{s}_t$ representations forms a richer 4D data stream (3D space + time). While $\hat{s}_t$ is fundamentally an enriched, structured \textit{observation}, it is referred to as \NickName's `state representation' for operational simplicity.

To mitigate drift from non-Markovian observations in POMDPs, \NickName employs an adaptive memory architecture, framing the task as a sequential auto-regressive prediction:
\begin{equation}
    \label{eq:prediction_model_obs}
    f_\theta = P\left(\hat{s}_{t+1:t+k} \mid a_t, \hat{s}_t, \hat{s}_{t-m:t-1}, \psi\left(\hat{s}_{0:t-m-1}\right)\right).
\end{equation}
The model $f_\theta$ in \NickName takes as inputs the action $a_t$, the current composite state representation $\hat{s}_t$, and the $m$ most recent past representations $\hat{s}_{t-m:t-1}$. A selective mechanism incorporates an old representation $\hat{s}_i = \psi(\hat{s}_{0:t-m-1})$ only if it is statistically significantly correlated with $\hat{s}_t$. 

%%%%%%%%%%%%%%%%%%%%%%%%%%%%%%%%%%%%%%%%%%%%%%%%%%%%%%%%%%%%%%%%%%%%%%%%%%%%%%%%%%%%%%%%%%%%%%%%%%%%%%%%%%%%%%%%
%%%%%%%%%%%%%%%%%%%%%%%%%%%%%%%%%%%%%%%%%%%%%%%%%%%%%%%%%%%%%%%%%%%%%%%%%%%%%%%%%%%%%%%%%%%%%%%%%%%%%%%%%%%%%%%%
%%%%%%%%%%%%%%%%%%%%%%%%%%%%%%%%%%%%%%%%%%%%%%%%%%%%%%%%%%%%%%%%%%%%%%%%%%%%%%%%%%%%%%%%%%%%%%%%%%%%%%%%%%%%%%%%
\subsection{Model Components}
\label{sec:Model Components}

{\noindent \textbf{4D Representation.}}
As formulated in Eq.~\ref{eq:4d_representation}, \NickName employs a 4D representation for state estimation.
Specifically, each $g$ is a tensor with dimensions matching the input image, where each pixel stores a 3D coordinate~\cite{wang2024dust3r}.
Following Aether~\cite{team2025aether}, we decompose 3D coordinates into depth $d_t$ and viewpoint components. This allows depth information to be directly encoded by the pre-trained Variational Autoencoder (VAE)~\cite{kingma2022vae, ke2024marigold}.
% This decomposition leverages the fact that depth information can be directly encoded by the pre-trained Variational Autoencoder (VAE) \cite{kingma2022vae, ke2024marigold} of existing image/video generation models, eliminating the need for additional VAE training.
Moreover, as described in Aether~\cite{team2025aether}, we parameterize depth values $d_t$ as the square root of disparity $e_t = \sqrt{1/d_t}$.
Finally, we adopt the raymap representation~\cite{team2025aether, chen2024and} to parameterize viewpoint $c$, which geometrically encodes camera orientation and position through ray direction vectors in the scene coordinate system.
We construct $\hat{s}$ by channel-wise concatenating the three modalities.
This unified structure ensures compatibility with standard image latents, enabling autoregressive future prediction through iterative generation.
%This unified 4D representation enables autoregressive future prediction via iterative generation - since our representation maintains structural compatibility with standard image latents, 
%we employ patchify procedures \cite{dosovitskiy2021vit} to decompose it into spatially organized tokens for neural network processing.
\begin{figure}[t]
    \begin{center}
    % \vspace{-0.5em}
        \includegraphics[width=\textwidth]{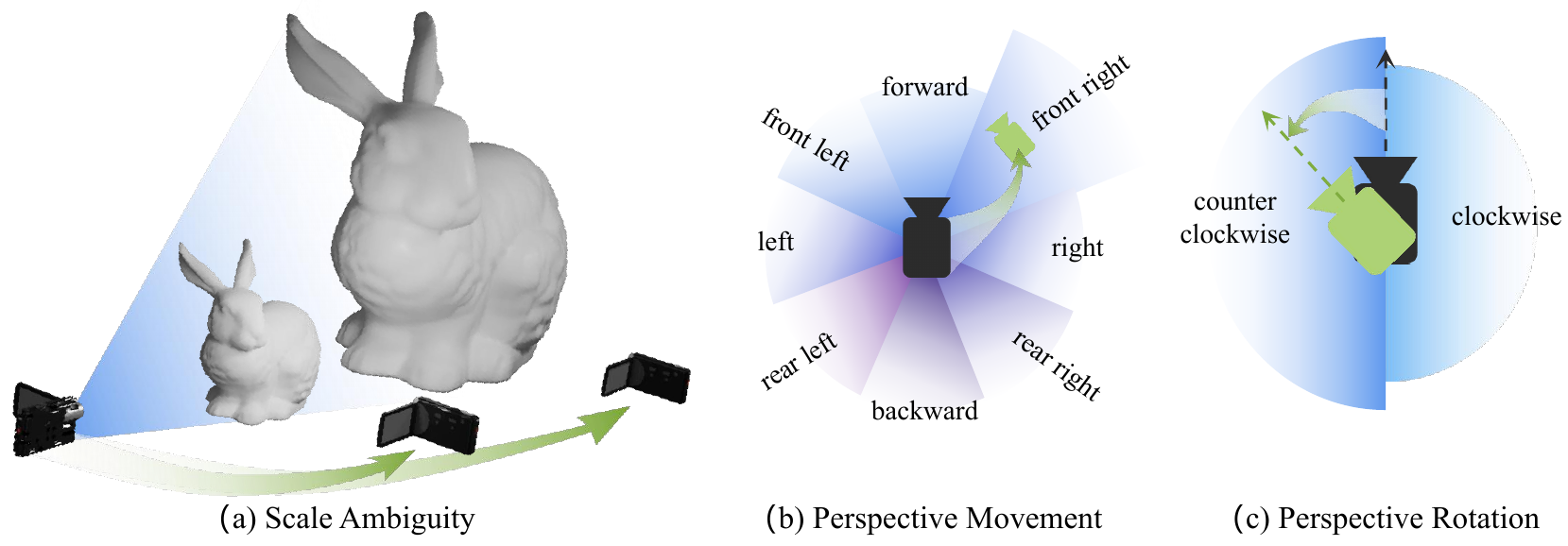}
        % \vspace{-2em}
        \caption{
            (a) Inferring 3D environments from a single image results in inherent scale ambiguity, a latent variable conditioned on the training data. Generating novel views from images alone, without 3D priors, is significantly more challenging than with explicit 3D structures, often leading to geometrically inconsistent extrapolations and error propagation in autoregressive predictions.
(b)(c) Text descriptions of perspective changes can be algorithmically derived from camera pose variations.
        }
        \label{fig:scale_camera}
    \end{center}
    % \vspace{-1.5em}
\end{figure}
% \paragraph{General Control}

{\noindent \textbf{General Control.}}
In previous studies \cite{feng2024thematrix,valevski2024gamengen}, controller data was concurrently collected during the data acquisition phase and subsequently integrated into model training as an additional modality. 
However, our \NickName\space framework deliberately avoids introducing new modalities, primarily for two reasons: First, to maximize preservation and utilization of the pre-trained model's capabilities; second, textual conditions inherently constitute a more versatile control paradigm. 
This design philosophy facilitates both direct text-to-controller key mapping in downstream applications and efficient fine-tuning on novel controllers. 
% 

% \paragraph{Spatial Condition}
{\noindent \textbf{Spatial Condition.}}
Since we explicitly model the 4D representation, we maintain a memory pool that stores all historical observations with their spatial positions aligned to the coordinate system of the initial observation. %, forming a unified reference framework. 
Through the camera pose in the current state, our selective mechanism $\psi$ dynamically retrieves a historical state as the spatial condition. 
This selected state is subsequently encoded into a token sequence to serve as the spatial condition. 
Our approach draws inspiration from the spatial neighbor state selection strategy employed in GigaGS \cite{chen2024gigags,gao2024cosurfgs}, where geometrically relevant historical states are prioritized based on spatial proximity:
\begin{equation}
    \psi(\hat{s}_{t} || \{\hat{s}_{t-1}, \cdots, \hat{s}_0\}) = \underset{j \in S}{\arg\min} \angle(R_t, R_j),\quad 
    \text{where}~ S = \underset{i \in \{t-1, \cdots, 0\}}{\arg\min^{(k)}}~(T_t - T_i)^2.
\end{equation}
% \begin{equation}
%     \psi(s_{n} || \{s_{n-1}, \cdots, s_0\}) =
%     \underset{j \in S}{\arg\min}  \angle(R_n, R_j) 
%     \underset{i \in S, |S| = k}{\arg\min}~(T_n - T_i)^2
% \end{equation}

% \begin{equation}
%     \psi(\cdot) = \underset{j \in \left( \underset{i \in I}{\arg\min^{(k)}}~(T_n - T_i)^2 \right)}{\arg\min} \angle(R_n, R_j)
% \end{equation}
Here, $R$ and $T$ denote the rotation and translation matrices of viewpoint $c$ in state $\hat{s}$.

%%%%%%%%%%%%%%%%%%%%%%%%%%%%%%%%%%%%%%%%%%%%%%%%%%%%%%%%%%%%%%%%%%%%%%%%%%%%%%%%%%%%%%%%%%%%%%%%%%%%%%%%%%%%%%%%
%%%%%%%%%%%%%%%%%%%%%%%%%%%%%%%%%%%%%%%%%%%%%%%%%%%%%%%%%%%%%%%%%%%%%%%%%%%%%%%%%%%%%%%%%%%%%%%%%%%%%%%%%%%%%%%%
%%%%%%%%%%%%%%%%%%%%%%%%%%%%%%%%%%%%%%%%%%%%%%%%%%%%%%%%%%%%%%%%%%%%%%%%%%%%%%%%%%%%%%%%%%%%%%%%%%%%%%%%%%%%%%%%
\subsection{Dataset Construction}
\label{sec:Dataset Construction}

% Initially, we collected approximately 10M frames of gameplay footage from AAA games, utilizing ReShade \cite{reshade} to systematically eliminate all UI elements. 
Initially, we collected approximately 10M frames of gameplay footage, utilizing ReShade \cite{reshade} to systematically eliminate all UI elements.
Subsequently, building upon the automatic camera annotation pipeline referenced in Aether \cite{team2025aether}, we synthesized datasets containing precise intrinsic/extrinsic camera parameters, depth maps, and high-fidelity synthetic images. 
The acquired camera parameters were first employed to filter out potentially contaminating data points. 
To facilitate interactive applications, we implemented hierarchical annotation protocols: textual labeling at the video clip level coupled with motion-specific labeling for finer-grained frame chunks.

% \paragraph{Filtering Criteria}
{\noindent \textbf{Filtering Criteria.}}
% In Model 2, we employ a 3D causal VAE for video encoding, which performs compression in both spatial and temporal dimensions. 
% Our analysis reveals that excessive camera rotation magnitudes or rapid view transitions significantly degrade the reconstruction quality after the VAE-based encoding-decoding process. 
Excessive camera rotation magnitudes or rapid view transitions significantly degrade the reconstruction quality after the 3D VAE-based encoding-decoding process. 
To address this, we establish a chunk-wise data filtering criterion: 
The chunk size is defined as the temporal compression ratio of the VAE. 
A video clip is considered valid only if the cumulative rotation angle across all chunks remains below a predefined threshold $\delta_{rot}$. 
The chunk-wise rotation angle is computed as the angular difference in forward direction between the last frames of each chunk. 
Additionally, we filter video clips exhibiting minimal camera/character movement by calculating displacement metrics through camera extrinsic parameters. 
Specifically, clips with movement distances (derived from extrinsic parameters) below a specified threshold $\delta_{move}$ are systematically excluded from the dataset.

% \paragraph{Caption Annotation}
{\noindent \textbf{Caption Annotation.}} 
% In previous studies \cite{feng2024thematrix,valevski2024gamengen}, controller data was concurrently collected during the data acquisition phase and subsequently integrated into model training as an additional modality. 
% However, our \NickName\space framework deliberately avoids introducing new modalities, primarily for two reasons: First, to maximize preservation and utilization of the pre-trained model's capabilities; second, textual conditions inherently constitute a more versatile control paradigm. 
% This design philosophy facilitates both direct text-to-controller key mapping in downstream applications and efficient fine-tuning on novel controllers. 
Given the precise positional data obtained, we initially construct textual descriptions directly from camera movements/rotations as illustrated in Figure \ref{fig:scale_camera}. 
Furthermore, we employ Qwen-VL \cite{Qwen-VL} to annotate video clips, generating first-person descriptions for viewpoint transitions in egocentric videos, while creating third-person narratives for character actions and movements in exocentric recordings.
We employ CLIP \cite{radford2021clip} and T5 \cite{raffel2023t5} to generate caption embeddings, adopting a methodology consistent with that implemented in SD3 \cite{esser2024sd3}.

% \paragraph{Training Preprocess}
{\noindent \textbf{Training Preprocess.}}  
For a video clip, global scene scaling based on scene dimensions is implemented to ensure effective compression. 
To guarantee that the depth values $d_i$ can be appropriately scaled into a constrained space for successful VAE encoding while preserving autoregressive causality, the entire sequence of $d_i$ is normalized by $d_{max} \times \lambda$, where $\lambda$ serves as a modulation factor. 
This normalization ensures that the $d$'s range of the initial frame is transformed into $(0, \lambda] \subseteq [0, 1] $, thereby reserving value space for subsequent frames where $d_i$ may exceed $d_{max}$. 
This mechanism effectively prevents truncation artifacts during the rescaling process to the VAE's input domain.

%%%%%%%%%%%%%%%%%%%%%%%%%%%%%%%%%%%%%%%%%%%%%%%%%%%%%%%%%%%%%%%%%%%%%%%%%%%%%%%%%%%%%%%%%%%%%%%%%%%%%%%%%%%%%%%%
%%%%%%%%%%%%%%%%%%%%%%%%%%%%%%%%%%%%%%%%%%%%%%%%%%%%%%%%%%%%%%%%%%%%%%%%%%%%%%%%%%%%%%%%%%%%%%%%%%%%%%%%%%%%%%%%
%%%%%%%%%%%%%%%%%%%%%%%%%%%%%%%%%%%%%%%%%%%%%%%%%%%%%%%%%%%%%%%%%%%%%%%%%%%%%%%%%%%%%%%%%%%%%%%%%%%%%%%%%%%%%%%%
\subsection{Long-Duration Inference}
\label{sec:Long-Duration Inference}

\begin{wrapfigure}{r}{0.5\textwidth} 
    \vspace{-1.5em}
    \begin{minipage}{0.5\textwidth}
    \begin{algorithm}[H]
        \caption{Long-Duration Inference}
        \label{alg: Long-Duration Inference}

        \SetKwProg{Fn}{Function}{}{}
        \SetKwInOut{Input}{Input}
        \SetKwInOut{Output}{Output}
    
        \Input{Observation $v_0$} 
        \Output{State sequence $\{\hat{s}_t\}_{t=1}^{\infty}$}

        Initialize memory  ${\Bbb M} \leftarrow \{\hat{s_0} = (v_0, \mathbf{0}, \mathbf{0})\}$ \;

        Initialize cache  ${\Bbb C} \leftarrow \emptyset$ \;

        \For{Inference loops $i=1,2,...$}{
        
            Read recent memories ${\Bbb C} \leftarrow Recent({\Bbb M})$ \;
            
            Scale cache ${\Bbb C}$ \;
            
            \While{$Size({\Bbb C}) < CacheMaxSize$}{
            
                Retrieve $\hat{s_{spatial}} \leftarrow \psi(\hat{s_{now}} || {\Bbb M} )$ \;
                
                Read action $a_{now}$ \;
                
                $\hat{s}_{next} = f_{\theta}(\hat{a}_{now}, {\Bbb C}, \hat{s}_{spatial}) $ \;

                Cache state ${\Bbb C} = {\Bbb C} \cup \{ \hat{s}_{next} \}$
            }

            Rescale cache ${\Bbb C}$ \;

            Update memories  ${\Bbb M} = {\Bbb M} \cup {\Bbb C}$
            
        }
        
    \end{algorithm}
  \end{minipage}
  \vspace{-1em}
\end{wrapfigure}

To enable long-duration reasoning, we employ a sliding window approach \cite{song2025historyguide,chen2024diffusionforcing}. 
Specifically, after obtaining the sequence $\hat{s}_{t:t+k}$, we utilize $\hat{s}_{t+k-m+1:t+k}$ as the conditioning context for subsequent window computations. 
Prior to this process, a scaling operation is applied to the transitional segment: Using $\hat{s}_{t+k-m+1}$ as the initial frame of the next window, we compute $d_{max}$ from $d_{t+k-m+1}$ to scale the parameters $d$ and $c$ within $\hat{s}_{t+k-m+2:t+k}$. This critical $d_{max}$ value is recorded to facilitate parameter rescaling upon completion of window computations, thereby enabling seamless sequence concatenation and preservation of global contextual information. 
The complete operational procedure is algorithmically formalized as described in the designated algorithm \ref{alg: Long-Duration Inference}.

\section{Experiments}

The proposed \NickName\space constitutes a diffusion model operating in an autoregressive paradigm, where the temporal generation process strictly adheres to an autoregressive framework implemented through flow matching methodology \cite{liu2022flow, albergo2022flow, lipman2022flow}. 
Extensive efforts were devoted to enabling the \NickName\space framework to function properly, supported by comprehensive ablation studies to validate the effectiveness of our proposed approach. 
As detailed in Section \ref{sec:Different Model Architectures}, we systematically investigated two distinct MM-DiT-based \cite{esser2024sd3} architectures for historical information integration. 
To substantiate the necessity of 4D modality introduction in \NickName, a comparative analysis was conducted in Section \ref{sec:Ablations}, demonstrating the critical advantages of incorporating this novel modality. 
Finally, Section \ref{sec:Simulation Quality} presents the superior performance of \NickName.
% accompanied by our empirical observations regarding its operational characteristics.

%%%%%%%%%%%%%%%%%%%%%%%%%%%%%%%%%%%%%%%%%%%%%%%%%%%%%%%%%%%%%%%%%%%%%%%%%%%%%%%%%%%%%%%%%%%%%%%%%%%%%%%%%%%%%%%%
%%%%%%%%%%%%%%%%%%%%%%%%%%%%%%%%%%%%%%%%%%%%%%%%%%%%%%%%%%%%%%%%%%%%%%%%%%%%%%%%%%%%%%%%%%%%%%%%%%%%%%%%%%%%%%%%
%%%%%%%%%%%%%%%%%%%%%%%%%%%%%%%%%%%%%%%%%%%%%%%%%%%%%%%%%%%%%%%%%%%%%%%%%%%%%%%%%%%%%%%%%%%%%%%%%%%%%%%%%%%%%%%%
\subsection{Different Model Architectures}
\label{sec:Different Model Architectures}

\begin{figure}[t]
    \begin{center}
    \vspace{-0.5em}
        \includegraphics[width=\textwidth]{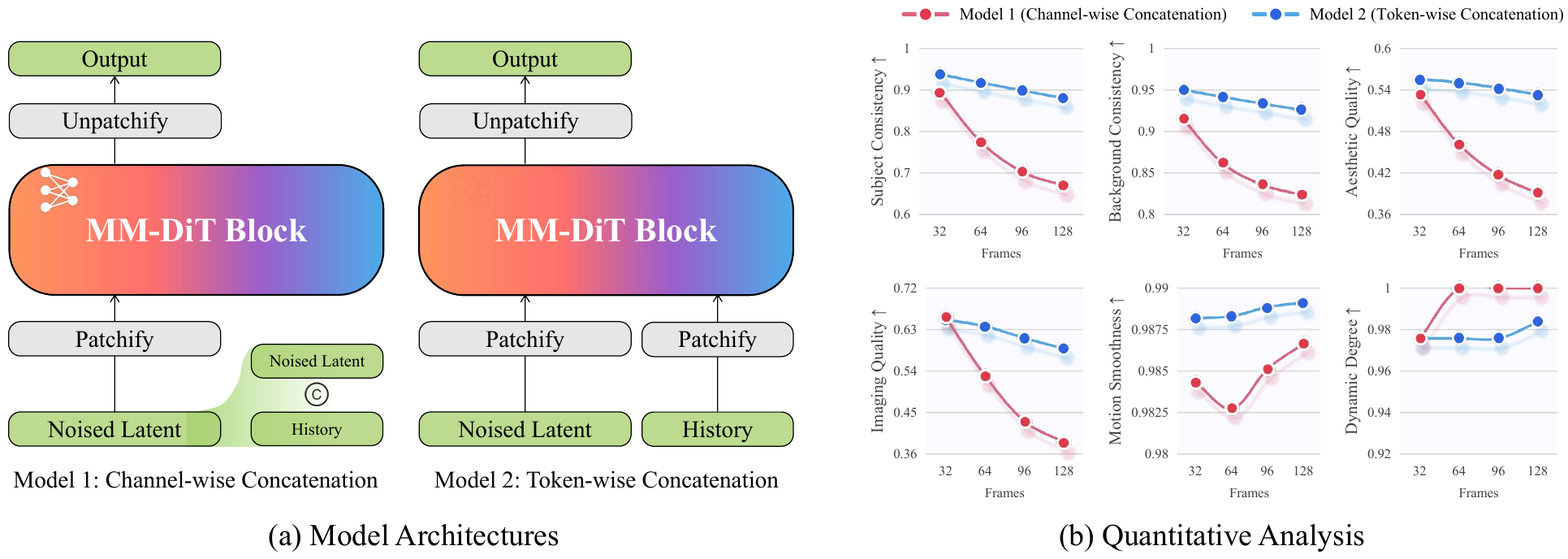}
        \caption{
            (a) Two MM-DiT-based architectures were designed to inject historical information.
            (b) Quantitative evaluation results on VBench \cite{huang2023vbench} demonstrate that Model 2 (Token-wise Concatenation) achieves superior performance in nearly all metrics, exhibiting enhanced visual quality and reduced temporal drift issues compared to alternative architectures.
        }
        \label{fig:ablation_arch_quan}
    \end{center}
    \vspace{-1.5em}
\end{figure}

In the training paradigm of diffusion models, historical and future observations are encoded into latent representations, which are subsequently patchified into tokens and input into a transformer-based \cite{vaswani2023transformer} network. 
As illustrated in figure \ref{fig:ablation_arch_quan} (a), \NickName\space explores two approaches for injecting historical information based on the MM-DiT \cite{esser2024sd3} architecture. We first adopt GameNGen's \cite{valevski2024gamengen} methodology by directly concatenating historical information through channel-wise concatenation. 
Subsequently, inspired by existing video generation methods \cite{jin2024pyramidalflow}, we develop a token-wise concatenation strategy to integrate temporal information.

% \paragraph{Model 1: Channel-wise Concatenation} 
{\noindent \textbf{Model 1: Channel-wise Concatenation.}}  
In this paradigm, the model initially employs an image encoding architecture to encode frames within a video clip into latent representations, where the temporal dimension of the latent space corresponds to the original video duration. 
Temporally ordered latent states from different timesteps are concatenated along the channel dimension. 
This concatenated representation is subsequently patchified into tokens for processing through the transformer architecture. 
Finally, these tokens are unpatchified and decoded into outputs with target channel dimensions. 
This methodology strategically avoids introducing additional tokens for future frames, instead integrating temporal information through concatenation operations that fuse noise tokens with historical states. 
Consequently, this design significantly reduces Floating Point Operations (FLOPs) per iteration, primarily attributed to the non-linear computational complexity inherent in transformer-based attention mechanisms.

% \paragraph{Model 2: Token-wise Concatenation}
{\noindent \textbf{Model 2: Token-wise Concatenation.}}  
In contrast to Model 1, this architecture demonstrates distinct characteristics in latent processing: After video encoding into latent representations, temporal states from different timesteps undergo individual patchification operations to generate different tokens. 
This approach substantially increases token quantity, necessitating the implementation of a 3D VAE framework to achieve temporal compression in the latent space. 
The resultant latent representations exhibit reduced dimensionality along the temporal axis compared to the original video clip's frame count, maintaining a fixed temporal compression rate 
(with the exception of the first frame) 
to balance information preservation and computational efficiency.

% \paragraph{Implement Details} 
{\noindent \textbf{Implementation Details.}}  
Both comparative models maintain identical parameter scales of 2 billion. 
To enhance training efficiency, we implemented Fully Sharded Data Parallelism (FSDP) \cite{zhao2023pytorchfsdp} with ZeRO-2 optimization for both architectures. 
For Model 1, parameter initialization was performed using pre-trained weights from the SD3-medium \cite{esser2024sd3}, whereas Model 2 utilized Pyramid-Flow \cite{jin2024pyramidalflow} initialization. 
Our experimental analysis revealed that excessive concatenation of historical frames in Model 1 failed to yield performance improvements, prompting adoption of a configuration with 7 historical frames and 1 noise frame. Moreover, we implement condition augmentation techniques \cite{ho2021cascadeddiffusion} on the historical frames.
Model 2 adheres to the Pyramid-Flow architecture's 57 frame protocol while implementing its dedicated 3D VAE for eightfold temporal compression. % across the sequence dimension.
All training videos underwent resolution standardization to $384p$ through bicubic interpolation, followed by center-cropping to achieve a uniform $4:3$ aspect ratio, with corresponding adjustments to camera intrinsic parameters in the metadata. 
Notably, Model 1 demonstrates reduced token counts per computational step compared to Model 2, enabling deployment of larger global batch sizes - specifically $512$ for Model 1 versus $256$ for Model 2. 
Both architectures employed the AdamW optimizer \cite{loshchilov2019adamw} with cosine annealing learning rate scheduling, incorporating linear warm up during the initial $1\%$ of training iterations.
All experiments were conducted on the NVIDIA A100 GPUs.

\paragraph{Quantitative Results} 
The evaluation conducted on VBench \cite{huang2023vbench} encompassed six metrics: \textit{subject consistency}, \textit{background consistency}, \textit{aesthetic quality}, \textit{imaging quality}, \textit{motion smoothness}, and \textit{dynamic degree}. 
Quantitative assessments were performed on the $32$, $64$, $96$, and $128$ generated frames, with comparative results graphically presented in figure \ref{fig:ablation_arch_quan}b. 
Our findings reveal that Model 2's token-wise concatenation mechanism, despite introducing higher computational complexity (quantified as average GFLOPs of $1280.9$ versus $1049.4$ for Model 1), effectively mitigates autoregressive model drift while achieving superior visual performance. 
Notably, while channel-wise concatenation demonstrated competitive performance \cite{valevski2024gamengen, alonso2024diamond} in specific applications such as the DOOM gaming environment, our analysis suggests that temporal feature aggregation within single tokens exacerbates error accumulation phenomena, particularly under extended scenarios. 
This empirical evidence substantiates our architectural preference for token-wise concatenation, which demonstrates enhanced robustness across temporal dimensions in large-scale multimodal domains.

%%%%%%%%%%%%%%%%%%%%%%%%%%%%%%%%%%%%%%%%%%%%%%%%%%%%%%%%%%%%%%%%%%%%%%%%%%%%%%%%%%%%%%%%%%%%%%%%%%%%%%%%%%%%%%%%
%%%%%%%%%%%%%%%%%%%%%%%%%%%%%%%%%%%%%%%%%%%%%%%%%%%%%%%%%%%%%%%%%%%%%%%%%%%%%%%%%%%%%%%%%%%%%%%%%%%%%%%%%%%%%%%%
%%%%%%%%%%%%%%%%%%%%%%%%%%%%%%%%%%%%%%%%%%%%%%%%%%%%%%%%%%%%%%%%%%%%%%%%%%%%%%%%%%%%%%%%%%%%%%%%%%%%%%%%%%%%%%%%
\subsection{Ablations}
\label{sec:Ablations}

As elaborated in Section \ref{sec:Different Model Architectures}, we have innovatively introduced a novel modality into the \NickName\space framework. 
In this section, we train an additional model that aligns with conventional autoregressive methodologies by excluding the depth modality, retaining solely the raymap-based camera representation. 
Notably, the experimental configuration maintains identical training methodologies, datasets, and initialization parameters for corresponding layers across all compared models. For quantitative evaluation, we adopt the FVD \cite{unterthiner2019fvd} and VBench \cite{huang2023vbench} as principal assessment criteria.

\begin{figure}[t]
    \begin{center}
    \vspace{-1em}
        \includegraphics[width=\textwidth]{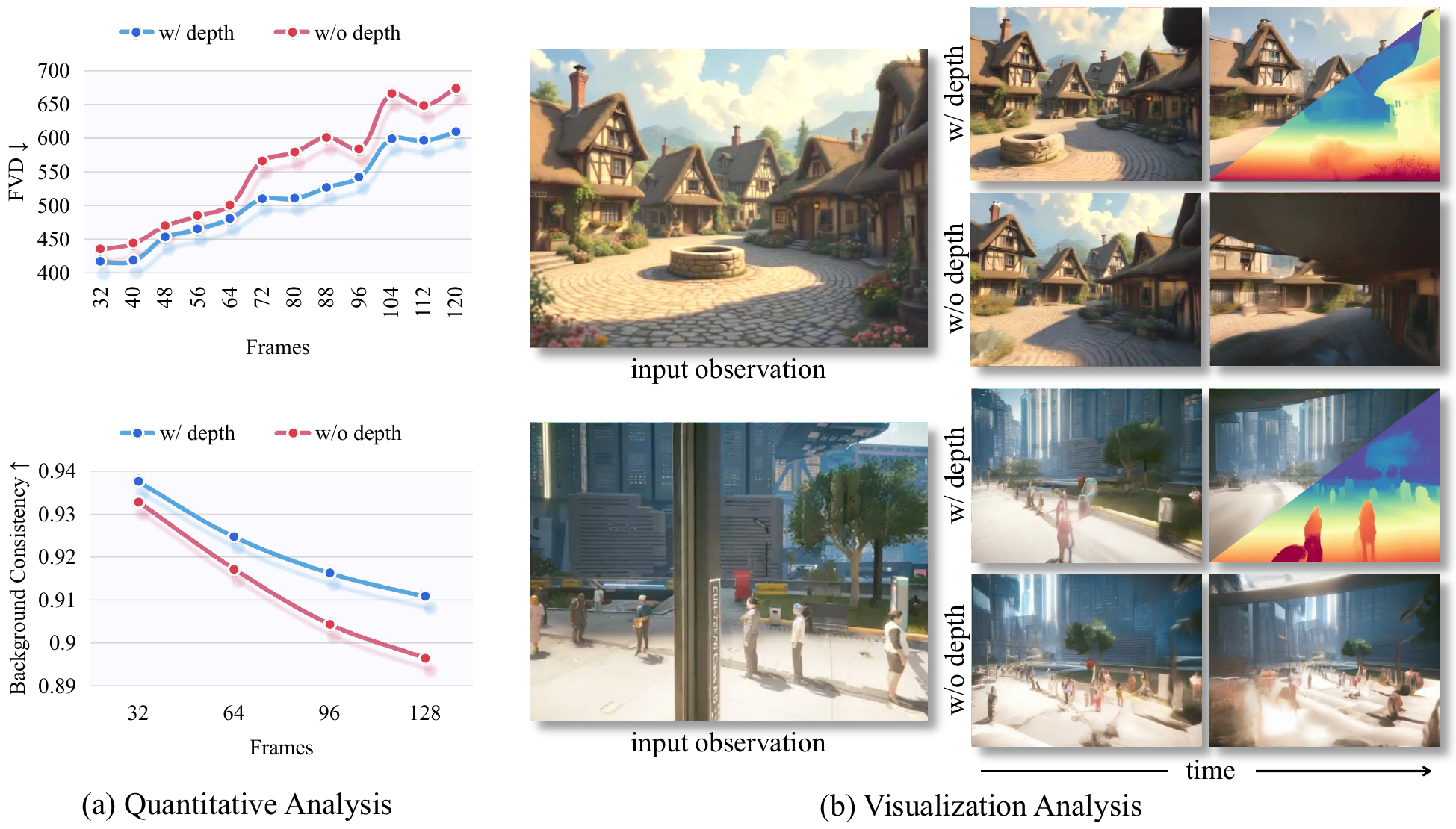}
        \caption{   
            \textbf{Ablation studies on depth modality.}
            (a) Quantitative results demonstrate that the integration of the depth modality yields superior performance in FVD and consistency.
            (b) Qualitatively, models incorporating depth exhibit enhanced environmental comprehension, achieving improved visual quality and mitigating temporal drift artifacts compared to the baseline.
        }
        \label{fig:ablation_depth}
    \end{center}
    \vspace{-1.5em}
\end{figure}

\begin{table}[!b]
    \centering
    \vspace{-1.5em}
    \caption{
        \textbf{Quantitative ablation study on depth modality.} Experiments conducted on six VBench metrics demonstrate that integrating depth modality achieves superior results, confirming the critical influence of 3D information on visual quality within autoregressive video generation frameworks.
    }
    \label{tab:ablation_vbench}
    
    \resizebox{\textwidth}{!}{
        \begin{tabular}{l|ccccccc}
            \toprule 
                        & \textit{frames} & \makecell{\textit{subject} \\ \textit{consistency}} & \makecell{\textit{background} \\ \textit{consistency}}  & \makecell{\textit{aesthetic} \\ \textit{quality}} &  \makecell{\textit{imaging} \\ \textit{quality}} & \makecell{\textit{motion} \\ \textit{smoothness}} & \makecell{\textit{dynamic} \\ \textit{degree}} \\
            \midrule
            \multirow{2}{*}{w/ depth (Ours)}  & 60   & \textbf{0.86939} & \textbf{0.92617} & \textbf{0.53415} & \textbf{0.48844} & \textbf{0.99032} & 1.00000 \\
                                              & 120  & \textbf{0.81652} & \textbf{0.91087} & \textbf{0.50028} & \textbf{0.44639} & \textbf{0.99147} & 1.00000 \\
            \multirow{2}{*}{w/o depth}        & 60   & 0.83602 & 0.91899 & 0.49106 & 0.43774 & 0.98975 & 1.00000 \\
                                              & 120  & 0.76812 & 0.89650 & 0.44095 & 0.37975 & 0.99062 & 1.00000 \\
            \bottomrule 
        \end{tabular}
    }
    \vspace{-1.5em}
\end{table}

% \paragraph{Introduction of New Modality}
{\noindent \textbf{Introduction of New Modality.}}  
We present a comparative visualization of two models in the figure \ref{fig:ablation_depth}, where both models generate predicted state sequences for future timesteps when initialized with a starting image and subjected to randomized action sequences. 
The empirical results demonstrate that the incorporation of depth modality substantially enhances the model's capacity to achieve comprehensive scene understanding, thereby enabling more precise estimation of latent world states that underlie observational inputs. 
This improved state estimation directly corresponds to enhanced visual predictive capabilities, as evidenced by our quantitative evaluation of synthesized video quality. 
While temporal drift persists as a fundamental challenge in autoregressive generation \cite{zhang2025framepack}, our findings reveal that depth integration effectively alleviates this deterioration phenomenon, with measurable improvements observed both in quantitative metrics and visual representations.

\begin{figure}[t]
    \begin{center}
        \includegraphics[width=\textwidth]{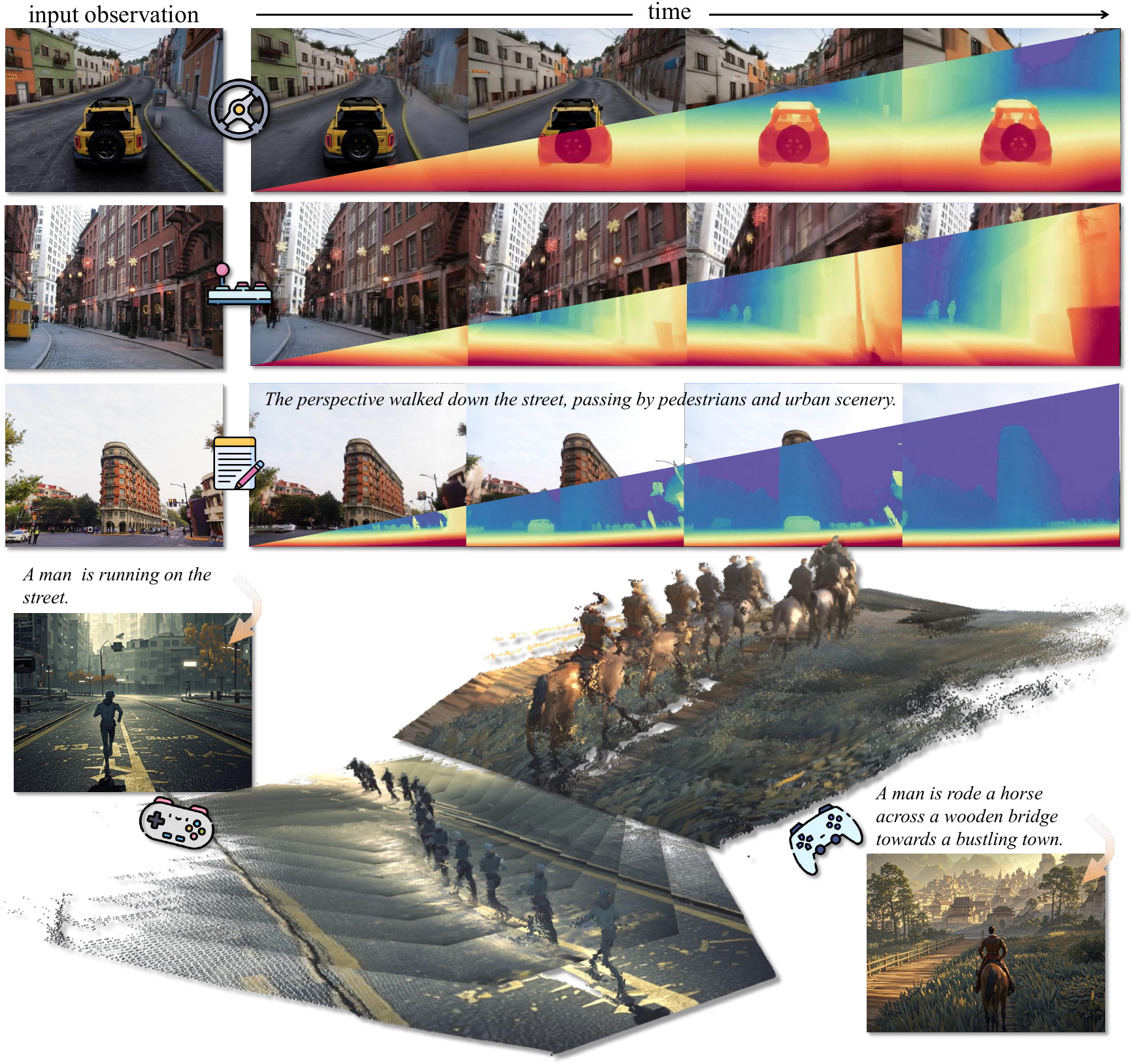}
        % \vspace{-2em}
        \caption{   
            Visualization Results.
        }
        \label{fig:results}
    \end{center}
    \vspace{-2.1em}
\end{figure}

\begin{figure}[t]
    \begin{center}
    \vspace{-0.5em}
        \includegraphics[width=\textwidth]{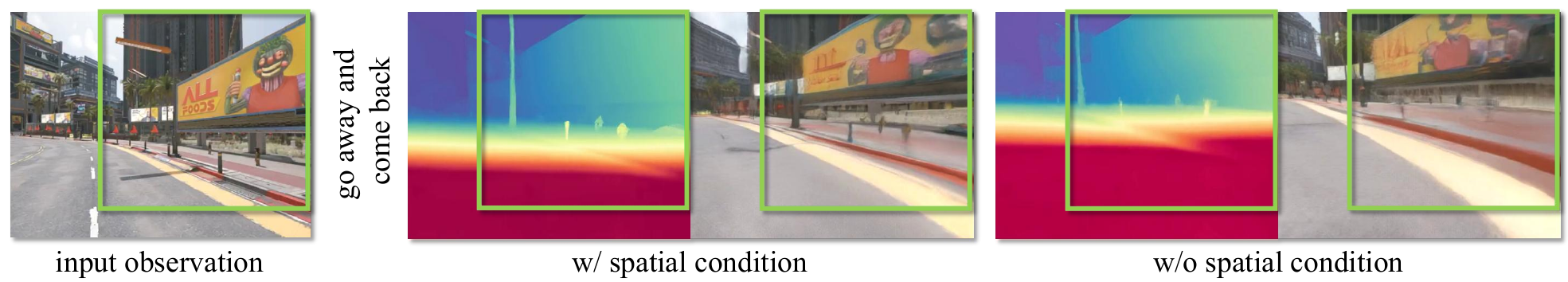}
        % \vspace{-2em}
        \caption{   
            The visualization of the spatial condition's effect.
        }
        \label{fig:ablation_spatial}
    \end{center}
    \vspace{-1.5em}
\end{figure}

% \paragraph{Spatial Memory}
{\noindent \textbf{Spatial Memory.}}  
By simultaneously predicting 3D camera poses during the generative process, we establish and maintain a global coordinate system anchored at the origin point defined by the initial frame's position. 
Our methodology implements a retrieval mechanism that queries the most recent pose from historical states to serve as spatial conditioning. 
During training, we strategically incorporate this spatial condition at controlled intervals as an additional modal constraint alongside textual inputs. For inference procedures, we adapt the InstructPix2Pix \cite{brooks2023instructpix2pix} framework for conditional generation. 
As demonstrated in the figure \ref{fig:ablation_spatial}, the integration of spatial conditioning enables extended temporal coherence in sequence generation that transcends the inherent limitations of fixed-duration video chunks, thereby achieving long-term spatial memory retention.

%%%%%%%%%%%%%%%%%%%%%%%%%%%%%%%%%%%%%%%%%%%%%%%%%%%%%%%%%%%%%%%%%%%%%%%%%%%%%%%%%%%%%%%%%%%%%%%%%%%%%%%%%%%%%%%%
%%%%%%%%%%%%%%%%%%%%%%%%%%%%%%%%%%%%%%%%%%%%%%%%%%%%%%%%%%%%%%%%%%%%%%%%%%%%%%%%%%%%%%%%%%%%%%%%%%%%%%%%%%%%%%%%
%%%%%%%%%%%%%%%%%%%%%%%%%%%%%%%%%%%%%%%%%%%%%%%%%%%%%%%%%%%%%%%%%%%%%%%%%%%%%%%%%%%%%%%%%%%%%%%%%%%%%%%%%%%%%%%%
\subsection{Simulation Quality}
\label{sec:Simulation Quality}

As illustrated in Figure \ref{fig:results}, we demonstrate the capabilities of \NickName\space through comprehensive evaluations. 
For each experimental instance, we exclusively employ visual inputs as initial observations, which include game images, real-world images, and AI-generated images (produced using the text-to-image model, Dreamina). % ~\cite{Dreamina}
Benefiting from our model's versatile conditioning mechanism, human-guided manipulations from various controllers can be manually projected into textual conditions for model input, while \NickName\space inherently supports direct textual condition integration. 
Our future prediction framework achieves highly consistent 4D representations while maintaining exceptional visual fidelity, with strict adherence to input conditions. 
Notably, the \NickName\space world model – grounded in 4D autoregressive video generation – distinguishes itself from conventional reconstruction-then-rerendering paradigms 
by simultaneously preserving viewpoint-object dynamics and predicting environmental interactions.

\section{Related Works}

{\noindent \textbf{Neural World Simulation.}} Neural world simulation employs generative models for dynamic, interactive environments, simulating real-world physics, a common limitation in standard video generation. 
UniSim~\cite{yang2024unisim} tackles this by training an action-conditioned video model with multi-dimensional datasets, creating an interactive universal simulator. UniPi~\cite{du2023unipi} reframes sequential decision-making as text-conditioned video generation, extracting control policies from generated future frames for cross-environment generalizability. Aether~\cite{team2025aether} argues videos are 2D projections and incorporates 3D structural information to better represent the underlying physical reality. Cosmos~\cite{nvidia2025cosmosworldfoundationmodel} shows that pre-training on physically-grounded video datasets, followed by fine-tuning, significantly enhances performance on physics-oriented AI tasks.

{\noindent \textbf{Interactive Video Generation.}} Interactive video generation merges interactivity with high-fidelity synthesis using neural networks.
Several approaches achieve controllable video generation by incorporating control operation labels into the generative training datasets: GameNGen~\cite{valevski2024gamengen}, Oasis~\cite{oasis2024}, DIAMOND~\cite{alonso2024diamond}, and GameFactory~\cite{yu2025gamefactory}.
Genie~\cite{bruce2024genie} introduces a Latent Action Model (LAM) to abstract generalized actions from extensive video data for universal control.
GameGen-X~\cite{che2024gamegenx} enables controllable video generation by pretraining on text-video pairs and then fine-tuning with other control modalities.
WorldMem~\cite{xiao2025worldmem} uses 3D pose representation for historical data retrieval to enhance long-term memory in video generation.

{\noindent \textbf{3D/4D Representations.}} 
The increasing integration of 3D and 4D representations~\cite{mildenhall2021nerf,zhu2023x,kerbl20233d,wu20244d,zhu2023ponderv2,yang2024unipad,wang2024dust3r,zhang2024cameras,he2025meshcraft,he2024gvgen} is proving transformative across multiple AI domains. In video generation, these higher-dimensional approaches are crucial for synthesizing dynamic scenes with enhanced spatial and temporal consistency~\cite{miao2025advances4dgenerationsurvey,yu20244realphotorealistic4dscene,lin2025exploringevolutionphysicscognition,jiang2025geo4d,zhang20244diffusionmultiviewvideodiffusion}.
World models benefit significantly from 3D/4D representations~\cite{zhen2025tesseract,team2025aether}, enabling a more profound understanding and prediction of environmental dynamics and the underlying physics governing them; these models strive to internalize spatial and temporal relationships to better simulate real-world scenarios. 
For embodied AI, 3D and 4D environmental awareness is fundamental~\cite{zhu2024spa,zhu2024point}, markedly improving agent capabilities in navigation~\cite{szot2021habitat} and manipulation~\cite{zhu2024spa,zhu2024point,lu2025h,xue2025demogen,ze20243d,fang2023rh20t,wang2024rise,yang2025fp3,jia2024lift3d}.
To the best of our knowledge, \NickName is the first to incorporate 4D representations into auto-regressive world models.

\section{Conclusion}

In this paper, we present \NickName, the first interactive world model based on 4D autoregressive video generation. 
We innovatively introduce 4D representation as our temporal observation to approximate the real world's environment. 
Our experimental results demonstrate the architectural effectiveness of the proposed model and quantitatively confirm the enhancements in visual quality and spatial capabilities achieved through the novel integration of 4D representation. 
Building upon this, we are capable of achieving long-duration inference and sustaining long-term memory capabilities.

{\noindent \textbf{Limitations.}}
Although \NickName has achieved promising results, its generalization capability to real-world scenarios remains limited due to being trained exclusively on synthetic data, necessitating further improvement in this aspect.

\begin{ack}

This work was done during Junyi Chen's internship at Shanghai AI Lab.
We thank Di Huang and Mingyu Liu for the valuable discussions.
This work is supported by the National Key R\&D Program of China (2022ZD0160102), and Shanghai Artificial Intelligence Laboratory.

\end{ack}

\bibliographystyle{plain}
\bibliography{ref}

%%%%%%%%%%%%%%%%%%%%%%%%%%%%%%%%%%%%%%%%%%%%%%%%%%%%%%%%%%%%

\newpage
\appendix

%%%%%%%%%%%%%%%%%%%%%%%%%%%%%%%%%%%%%%%%%%%%%%%%%%%%%%%%%%%%%%%%%%%%%%%%%%%%%%%%%%%%%%%%%%
%%%%%%%%%%%%%%%%%%%%%%%%%%%%%%%%%%%%%%%%%%%%%%%%%%%%%%%%%%%%%%%%%%%%%%%%%%%%%%%%%%%%%%%%%%
%%%%%%%%%%%%%%%%%%%%%%%%%%%%%%%%%%%%%%%%%%%%%%%%%%%%%%%%%%%%%%%%%%%%%%%%%%%%%%%%%%%%%%%%%%
\section{Training Details}

\subsection{Final Architecture}

\begin{figure}[h]
    \begin{center}
        \includegraphics[width=0.8\textwidth]{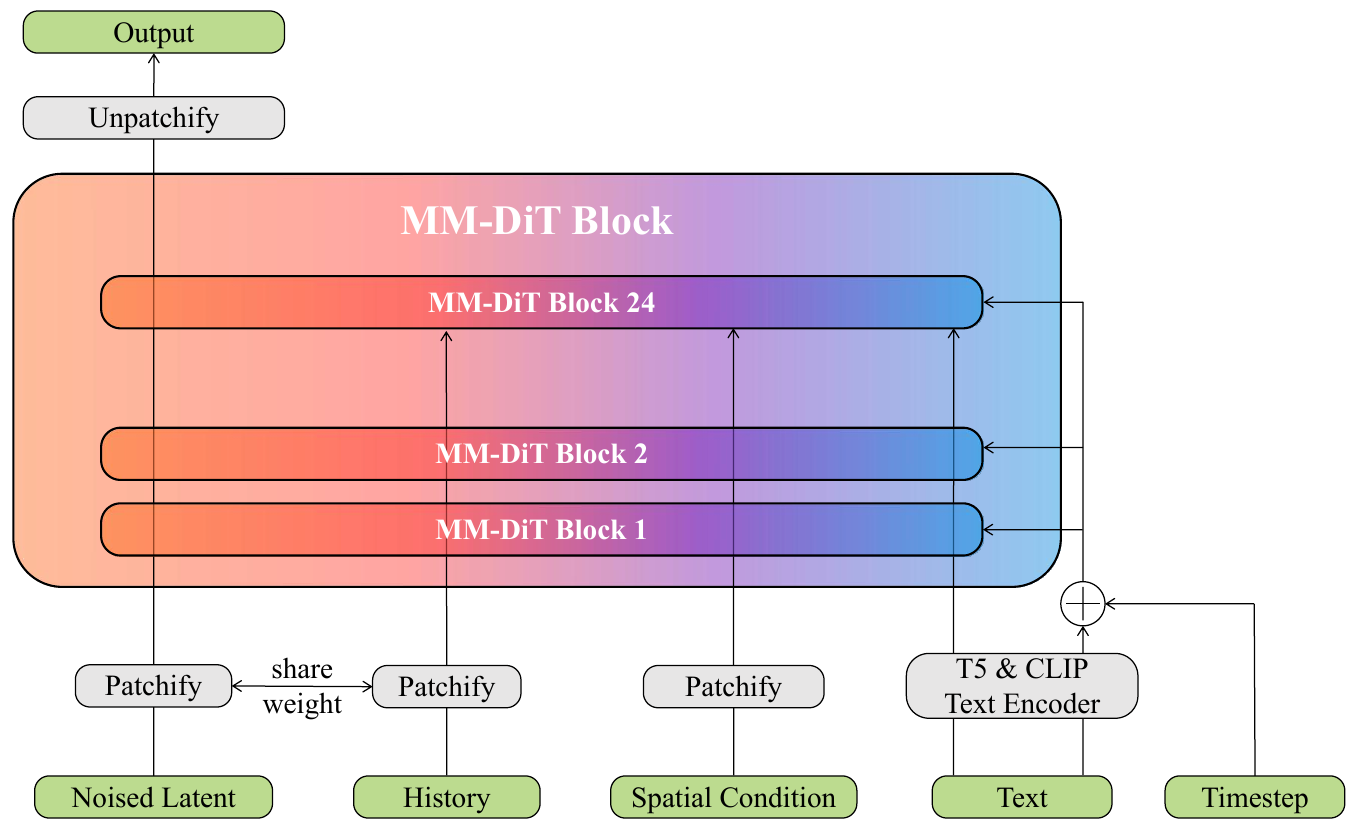}
        % \vspace{-1.5em}
        \caption{
            Final architecture.
        }
        \label{fig:arch_detail}
    \end{center}
    \vspace{-1em}
\end{figure}

Our final architecture employs a token-wise concatenation method. 
As illustrated in figure \ref{fig:arch_detail}, the noised latent and recent history undergo identical patchify operations to be encoded into tokens, which are subsequently concatenated along the token sequence dimension. 
Additionally, the spatial condition is independently processed through a patchify operation for token encoding. 
In alignment with SD3 \cite{esser2024sd3}, our text encoder integrates both T5 \cite{raffel2023t5} and CLIP \cite{radford2021clip} frameworks, with the obtained embeddings and pooled embeddings being injected into the model through token-wise concatenation and AdaLN mechanisms respectively. 
To ensure training stability, the MM-DiT \cite{esser2024sd3} architecture incorporates RMSNorm \cite{zhang2019rmsnorm} for QK Normalization \cite{henry2020qknorm}. 
The final model outputs are transformed via a linear projection layer to reconstruct tensors matching the shape of the original noise latent.
We present the parameters of our model in table \ref{tab:architecture para}.

\begin{table}[h]
    \centering
    % \vspace{-0.5em}
    \caption{
        Architecture parameters.
    }
    \label{tab:architecture para}
    \begin{tabular}{cc}
        \toprule
         layers & 24 \\
         model dimension & 1536 \\
         attention heads & 24 \\
         head dimension  & 64 \\
         spatial position embedding & sincos \\
         temporal position embedding & RoPE \cite{su2023rope} \\
         patch size & $2 \times 2$ \\
        \bottomrule 
    \end{tabular}
\end{table}

%%%%%%%%%%%%%%%%%%%%%%%%%%%%%%%%%%%%%%%%%%%%%%%%%%%%%%%%%%%%%%%%%%%%%%%%%%%%%%%%%%%%%%%%%%
\subsection{Raymap}

The raymap serves as an over-parameterized encoding mechanism for 3D viewing, generated through a ray-casting process where each pixel in the image plane emits a directional ray originating from the camera's optical center. 
This representation maintains spatial correspondence with the original image dimensions while containing $6$ channels of geometric information: $3$ channels encode the ray origin coordinates (equivalent to the camera position in 3D space), and the remaining $3$  channels specify the unit direction vectors of each cast ray. 
Notably, this parametrization preserves sufficient geometric constraints to enable camera parameter recovery through the reconstruction algorithm \ref{alg:raymap-to-cam-pose}.

\begin{wrapfigure}{r}{0.6\textwidth}
    \vspace{-1.5em}
    \begin{minipage}{0.6\textwidth}
    \begin{algorithm}[H] 
        \caption{Raymap to camera parameters conversion.}
        \label{alg:raymap-to-cam-pose}
        
        \SetKwProg{Fn}{Function}{}{}
        \SetKwInOut{Input}{Input}
        \SetKwInOut{Output}{Output}
    
        \Input{Raymap $c$} 
        \Output{Camera Parameters $intrinsic, extrinsic$}

        Estimate camera position $ T \leftarrow Ray\_o (Raymap)$
        
        Estimate ray directions $ Ray\_d \leftarrow Ray\_d (Raymap)$

        Calculate $intrinsic$ from $Ray\_d$

        Calculate camera rotation $R$ from $Ray\_d$

        $extrinsic \leftarrow R, T$
        
        return $intrinsic, extrinsic$
        
    \end{algorithm}
  \end{minipage}
  \vspace{-1em}
\end{wrapfigure}

%%%%%%%%%%%%%%%%%%%%%%%%%%%%%%%%%%%%%%%%%%%%%%%%%%%%%%%%%%%%%%%%%%%%%%%%%%%%%%%%%%%%%%%%%%
\subsection{Compact 4D Representation}

The final 3D VAE architecture adopted in \NickName achieves a temporal compression ratio of $8$ along the sequence dimension, enabling the prediction of consecutive future observations spanning $8$ time steps. 
While both image and depth modalities are encoded with $16$ latent channels through the 3D VAE, the raymap modality resists effective compression via this architecture. 
To address this, raymap data undergoes spatial downsampling through average pooling to match the latent dimensions of the image modality, followed by temporal concatenation. 
This configuration results in a combined channel count of $80$ ($16 + 16 + 6 \times 8$), with the majority allocated to raymap representation. 
However, considering the primary learning challenges reside in the image and depth modalities, we implement a keyframe optimization strategy: 
Only the final observation in each $8$-step sequence is retained as the keyframe, with its complete raymap concatenated ($6$ channels), while intermediate frames' raymaps are generated through linear interpolation of adjacent keyframes. This approach reduces the input dimensionality to $38$ channels ($16 + 16 + 6$) while maintaining temporal coherence. 
The methodological validity stems from two key observations: 1) Construction of globally consistent 4D representations requires only keyframe inclusion rather than full-sequence encoding, and 2) This selective encoding significantly reduces both global memory requirements and model input complexity, particularly beneficial for maintaining computational efficiency in long-term sequence modeling.

%%%%%%%%%%%%%%%%%%%%%%%%%%%%%%%%%%%%%%%%%%%%%%%%%%%%%%%%%%%%%%%%%%%%%%%%%%%%%%%%%%%%%%%%%%
\subsection{Data Batch}

Through data annotation and filtering protocols, we partition all video content into approximately 30,000 non-overlapping video splits, with each split constrained to a maximum of 400 frames. 
Under our training configuration, we sample $b$ consecutive $57$-frame video clips as a single batch. 
The system pre-specifies the potential quantity of video clips contained within each split. 
Notably, while video splits maintain non-overlapping boundaries, individual clips within the same split may exhibit temporal overlap. 
This methodology ultimately yields a curated dataset of $1.5$ million video clips. 
For enhanced stability during autoregressive training, we implement a GPU partitioning strategy where devices are grouped into clusters of size $8$ – this configuration precisely corresponds to the temporal dimension length of latent representations generated by 3D-VAE processing of $57$-frame sequences.
Within each group, GPUs are assigned identical input batches but process distinct temporal target segments.

%%%%%%%%%%%%%%%%%%%%%%%%%%%%%%%%%%%%%%%%%%%%%%%%%%%%%%%%%%%%%%%%%%%%%%%%%%%%%%%%%%%%%%%%%%
\subsection{Training Target}

\begin{figure}[h]
    \begin{center}
        \includegraphics[width=0.7\textwidth]{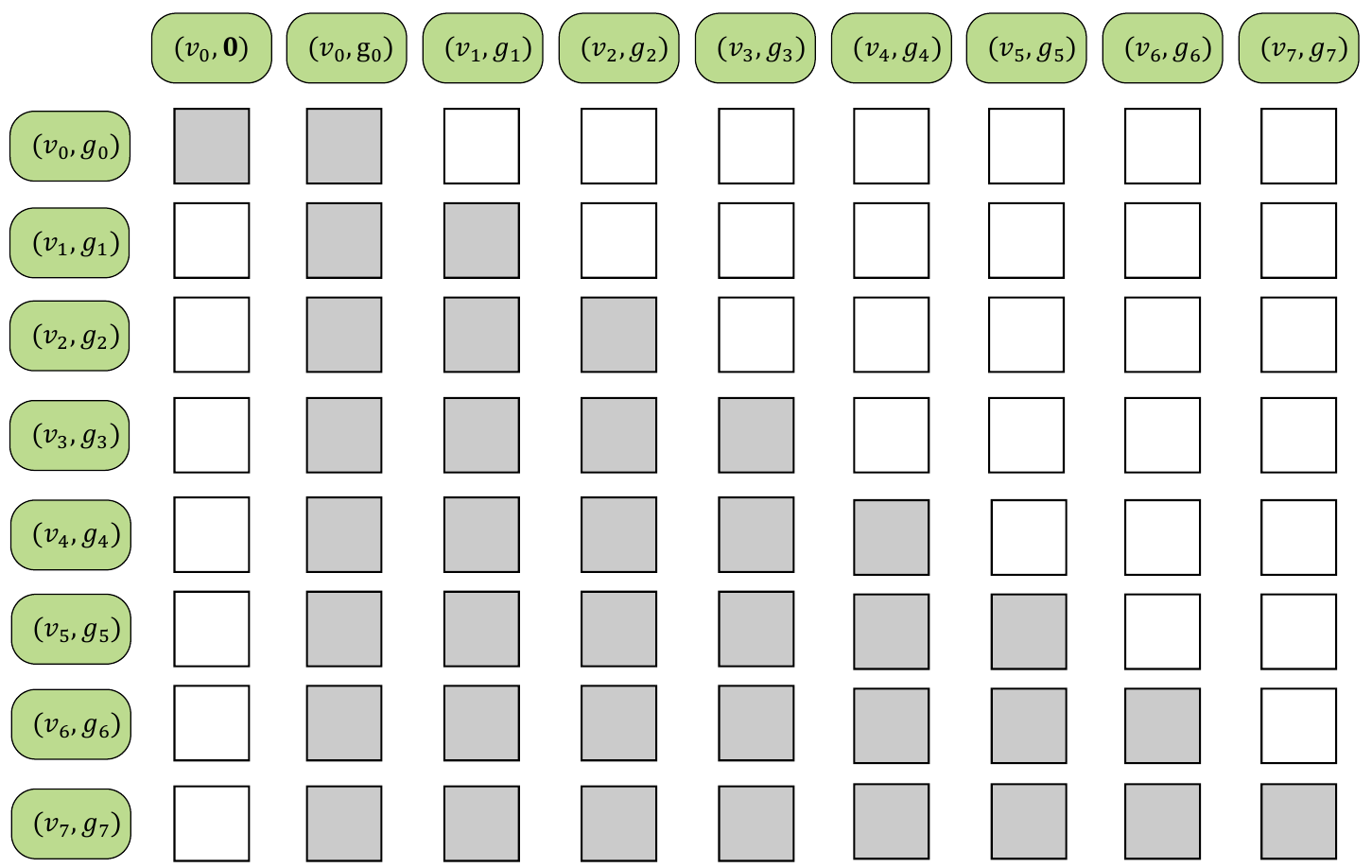}
        % \vspace{-1.5em}
        \caption{
            Attention mask during training.
        }
        \label{fig:attention_mask}
    \end{center}
    \vspace{-1em}
\end{figure}

\NickName can predict future 4D representations at various subsequent timesteps utilizing only a single input image $v_0$. Since the input image does not constitute a complete 4D representation, we first complete the 4D representation for that timestep. 
As illustrated in figure \ref{fig:attention_mask}, we initially predict the complete 4D representation $(v_0, g_0)$ corresponding to the input image, then employ this complete representation to replace the previously incomplete 4D representation. 
This procedure aligns with and maintains consistency with the inference phase.

{\noindent \textbf{Classifier-Free Guidance \cite{ho2022classifierfreediffusionguidance}.}} 
Our framework comprises two distinct condition components: textual condition and spatial condition. 
During the training phase, we employ stochastic conditioning dropout by masking textual condition $c_T$ with a 10\% probability and spatial condition $c_S$ with a 50\% probability. 
For inference, we implement the multimodal conditioning strategy with classifier-free guidance as proposed in InstructPix2Pix \cite{brooks2023instructpix2pix}, which coordinates the conditional fusion through learned guidance scales for each modality:

\begin{equation}
\begin{split}
    e_{\theta}(zt, c_T , c_S ) = & e_{\theta}\big(z_t \varnothing, \varnothing) \\ 
    &+ s_T \times \big(e_{\theta}(z_t, c_T , \varnothing) - e_{\theta}(z_t, \varnothing, \varnothing)\big) \\
    &+ s_S \times \big(e_{\theta}(z_t, c_T , c_S ) - e_{\theta}(z_t, c_T , \varnothing)\big).
\end{split}
\end{equation}

During the inference phase, we employ modality-specific guidance scales of $4$ and $5$ for textual and spatial condition respectively.

%%%%%%%%%%%%%%%%%%%%%%%%%%%%%%%%%%%%%%%%%%%%%%%%%%%%%%%%%%%%%%%%%%%%%%%%%%%%%%%%%%%%%%%%%%
\subsection{Training Resource}

To enhance training efficiency, we precomputed and stored text embeddings generated by T5 \cite{raffel2023t5} and CLIP \cite{radford2021clip} models, thereby eliminating the need to reload these text encoders or reprocess textual inputs during training. 
The entire training procedure spanned $2$ epochs, with our final model requiring approximately $23,000$ A100 GPU hours for completion.

%%%%%%%%%%%%%%%%%%%%%%%%%%%%%%%%%%%%%%%%%%%%%%%%%%%%%%%%%%%%%%%%%%%%%%%%%%%%%%%%%%%%%%%%%%
%%%%%%%%%%%%%%%%%%%%%%%%%%%%%%%%%%%%%%%%%%%%%%%%%%%%%%%%%%%%%%%%%%%%%%%%%%%%%%%%%%%%%%%%%%
%%%%%%%%%%%%%%%%%%%%%%%%%%%%%%%%%%%%%%%%%%%%%%%%%%%%%%%%%%%%%%%%%%%%%%%%%%%%%%%%%%%%%%%%%%
\section{Experiments Details}

\subsection{Metrics}

{\noindent \textbf{Fréchet Video Distance (FVD) \cite{unterthiner2019fvd}.}} 
The FVD is a metric used to evaluate the quality of generated videos by measuring the similarity between the distribution of real videos and synthesized videos. 
It leverages deep features extracted from pre-trained video models to compute the distance between real and generated video distributions in a high-dimensional feature space:

\begin{equation}
    \text{FVD} = \|\mu_r - \mu_g\|^2 + \text{Tr}(\Sigma_r + \Sigma_g - 2(\Sigma_r \Sigma_g)^{1/2})
\end{equation}

where $\mu_r, \mu_g$ are the mean vectors, and $\Sigma_r, \Sigma_g$ are the covariance matrices of real and generated video features.
In this paper, we employ I3D networks \cite{carreira2018i3d} pre-trained on the RGB frame data from the Kinetics-400 dataset \cite{kay2017kinetics} as our feature extraction framework.

{\noindent \textbf{VBench \cite{huang2023vbench}.}} 
VBench serves as a comprehensive evaluation benchmark suite for video generation models, designed to perform systematic assessments. 
This framework leverages a hierarchical evaluation structure that decomposes the multifaceted concept of "video generation quality" into well-defined constituent dimensions. 
In this paper, we adopt the six evaluation criteria: \textit{subject consistency}, \textit{background consistency}, \textit{aesthetic quality}, \textit{imaging quality}, \textit{motion smoothness}, and \textit{dynamic degree}, as our primary performance metrics.

%%%%%%%%%%%%%%%%%%%%%%%%%%%%%%%%%%%%%%%%%%%%%%%%%%%%%%%%%%%%%%%%%%%%%%%%%%%%%%%%%%%%%%%%%%
\subsection{Long-Duration Inference}

\begin{figure}[h]
    \begin{center}
        \includegraphics[width=\textwidth]{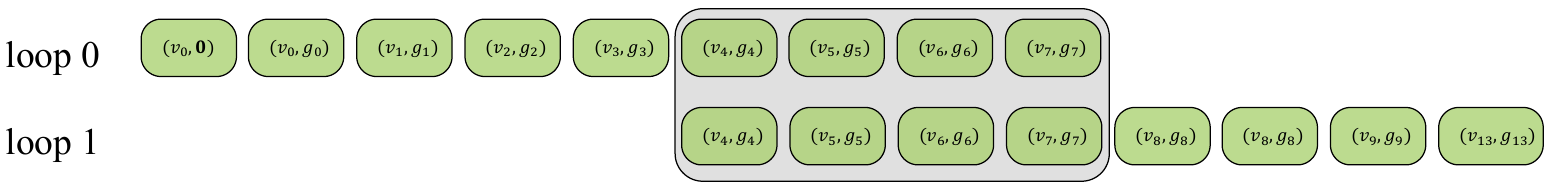}
        % \vspace{-1.5em}
        \caption{
            Long-duration inference.
        }
        \label{fig:long_inference}
    \end{center}
    % \vspace{-1em}
\end{figure}

During the training phase, we have modeled the distribution:
\begin{equation}
    P\left(\hat{s}_{t+1:t+k} \mid a_t, \hat{s}_t, \hat{s}_{t-m:t-1}, \psi\left(\hat{s}_{0:t-m-1}\right)\right). 
\end{equation}
To achieve long-duration inference, when the number of cached observations reaches the predefined \textit{CacheMaxSize} (set as the maximum video clip length during training), we first rescale all cached observations using the preserved $d_{max}$ parameters. 
These rescaled observations are then aligned to the global coordinate system through predicted camera parameters and stored in the memory. 
Subsequently, the most recent $m$ observations are directly adopted as the recent history, as shown in figure \ref{fig:long_inference}.
The first observation's $d_{max}$ value within this $m$-length sequence is utilized to scale these $m$ observations, followed by cache updating and subsequent predictions. 
This methodology ensures global consistency while enabling effective long-duration inference.

%%%%%%%%%%%%%%%%%%%%%%%%%%%%%%%%%%%%%%%%%%%%%%%%%%%%%%%%%%%%%%%%%%%%%%%%%%%%%%%%%%%%%%%%%%

%%%%%%%%%%%%%%%%%%%%%%%%%%%%%%%%%%%%%%%%%%%%%%%%%%%%%%%%%%%%

\end{document}